\documentclass[sigconf]{acmart}
\usepackage{graphicx}
\usepackage{amsmath}
\usepackage{booktabs}
\usepackage[ruled,linesnumbered]{algorithm2e}
\usepackage{subfigure}
\usepackage{threeparttable}
\usepackage{multirow}
\usepackage{arydshln}
\usepackage{makecell}
\usepackage{soul}
\usepackage{tabularx}
\usepackage{balance}
\usepackage{color}
\usepackage{bbding}
\usepackage{enumitem}
\usepackage{colortbl}

\AtBeginDocument{%
  \providecommand\BibTeX{{%
    \normalfont B\kern-.5em{\scshape i\kern-.25em b}\kern-.8em\TeX}}}

\setcopyright{acmlicensed}
\copyrightyear{2018}
\acmYear{2018}
\acmDOI{XXXXXXX.XXXXXXX}

\theoremstyle{definition}

\begin{document}
\newsavebox\CBox
\def\textBF#1{\sbox\CBox{#1}\resizebox{\wd\CBox}{\ht\CBox}{\textbf{#1}}}
\title{Mixed Blessing: Class-Wise Embedding guided Instance-Dependent Partial Label Learning}

\author{Fuchao Yang}
\email{yangfc@seu.edu.cn}
\affiliation{
  \institution{College of Software Engineering, Southeast University}
  \city{Nanjing}
  \country{China}
}

\author{Jianhong Cheng}
\email{chengjh@seu.edu.cn}
\affiliation{
  \institution{School of Computer Science and Engineering, Southeast University}
  \city{Nanjing}
  \country{China}
}

\author{Hui Liu}
\email{h2liu@sfu.edu.hk}
\affiliation{
  \institution{Yam Pak Charitable Foundation School of Computing and Information Sciences, Saint Francis University}
  \city{Hong Kong}
  \country{China}
}

\author{Yongqiang Dong}
\email{dongyq@seu.edu.cn}
\affiliation{
  \institution{School of Computer Science and Engineering, Southeast University}
  \city{Nanjing}
  \country{China}
}

\author{Yuheng Jia}
\email{yhjia@seu.edu.cn}
\authornote{Corresponding Author}
\affiliation{
  \institution{School of Computer Science and Engineering, Southeast University}
  \city{Nanjing}
  \country{China}
}

\author{Junhui Hou}
\email{jh.hou@cityu.edu.hk}
\affiliation{
  \institution{Department of Computer Science, City University of Hong Kong}
  \city{Hong Kong}
  \country{China}
}


\begin{abstract}
In partial label learning (PLL), every sample is associated with a candidate label set comprising the ground-truth label and several noisy labels. The conventional PLL assumes the noisy labels are randomly generated (instance-independent), while in practical scenarios, the noisy labels are always instance-dependent and are highly related to the sample features, leading to the instance-dependent partial label learning (IDPLL) problem. Instance-dependent noisy label is a double-edged sword. On one side, it may promote model training as the noisy labels can depict the sample to some extent. On the other side, it brings high label ambiguity as the noisy labels are quite undistinguishable from the ground-truth label. To leverage the nuances of IDPLL effectively, for the first time we create class-wise embeddings for each sample, which allow us to explore the relationship of instance-dependent noisy labels, i.e., the class-wise embeddings in the candidate label set should have high similarity, while the class-wise embeddings between the candidate label set and the non-candidate label set should have high dissimilarity. Moreover, to reduce the high label ambiguity, we introduce the concept of class prototypes containing global feature information to disambiguate the candidate label set. Extensive experimental comparisons with twelve methods on six benchmark data sets, including four fine-grained data sets, demonstrate the effectiveness of the proposed method.  The code implementation
is publicly available at \url{https://github.com/Yangfc-ML/CEL}.
\end{abstract}





\maketitle

\begin{figure}[ht]
  \centering
  \includegraphics[width=1\linewidth]{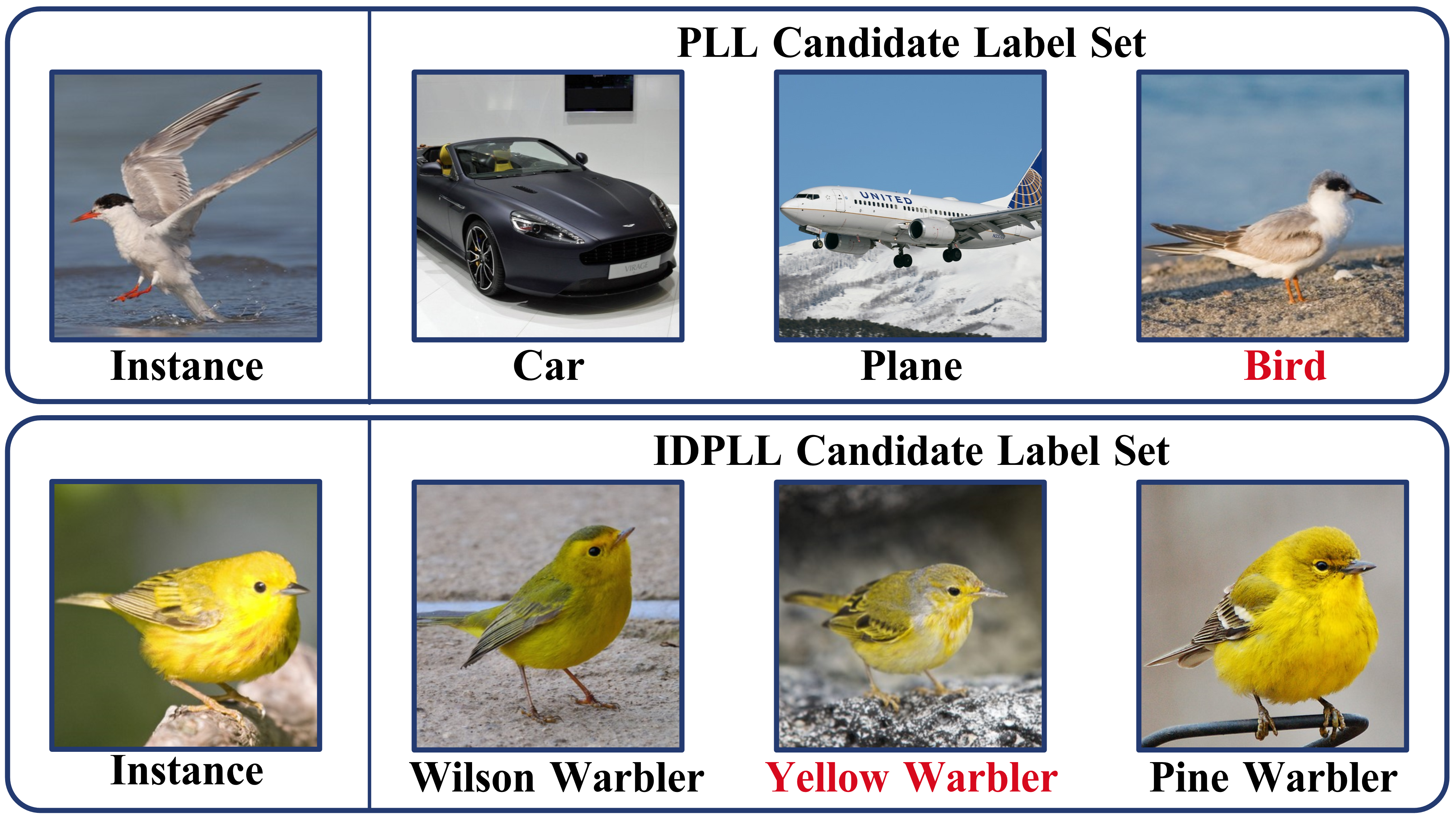}
  \caption{Differences between the conventional PLL and IDPLL, where the red label is the ground-truth label of the instance. In PLL, the noisy labels in the candidate label set are randomly generated. However, in IDPLL, the noisy labels in the candidate label set are instance-dependent, making them very similar to that of the ground-truth label, which brings more label ambiguity.}
  \label{sample}
\end{figure}

\section{Introduction}

Partial label learning (PLL) \cite{PLL1, PLL2, jiang, peng} has garnered significant attention over the past decade as a form of weakly supervised learning. In PLL, each sample is associated with a candidate label set, concealing the ground-truth label and several noisy labels within it. In the training phase, the ground-truth label is inaccessible. PLL does not rely on precise labeling, leading to substantial reductions in time and resource costs associated with sample annotation. Consequently, PLL finds extensive applications across various real-world domains, including automatic image annotation \cite{CLPL}, web mining \cite{Learning_from_Candidate_Labeling_Sets}, ecoinformatics \cite{MSRCv2}, and multimedia content analysis \cite{Soccer_Player}. The crux of addressing the PLL problem lies in label disambiguation, involving the identification of the ground-truth label while mitigating the impact of noisy labels within the candidate label set. Conventional label disambiguation techniques \cite{PRODEN, PICO, DPLL} have demonstrated strong efficacy in the typical PLL scenarios where noisy labels are randomly generated (instance-independent) \cite{AGGD,DPCLS}.

Recently, a more realistic PLL framework called instance-dependent partial label learning (IDPLL) \cite{VALEN,NEPLL,DIRK} was proposed, where noisy labels are re-tailored to individual samples (instance-dependent). As shown in Fig. \ref{sample}, in the conventional PLL, the noisy labels in the candidate label set may have significant differences with the ground-truth label because the noisy labels are randomly generated. However, in IDPLL, the noisy labels are very similar to the ground-truth label. To be more specific, these birds all have yellow feathers, which is more easily to cause label ambiguity. Conventional PLL methods often struggle in IDPLL since they ignore the characteristics of IDPLL. 

\begin{figure}[t]
  \centering
  \subfigure[CUB200]{
  \includegraphics[scale=0.29]{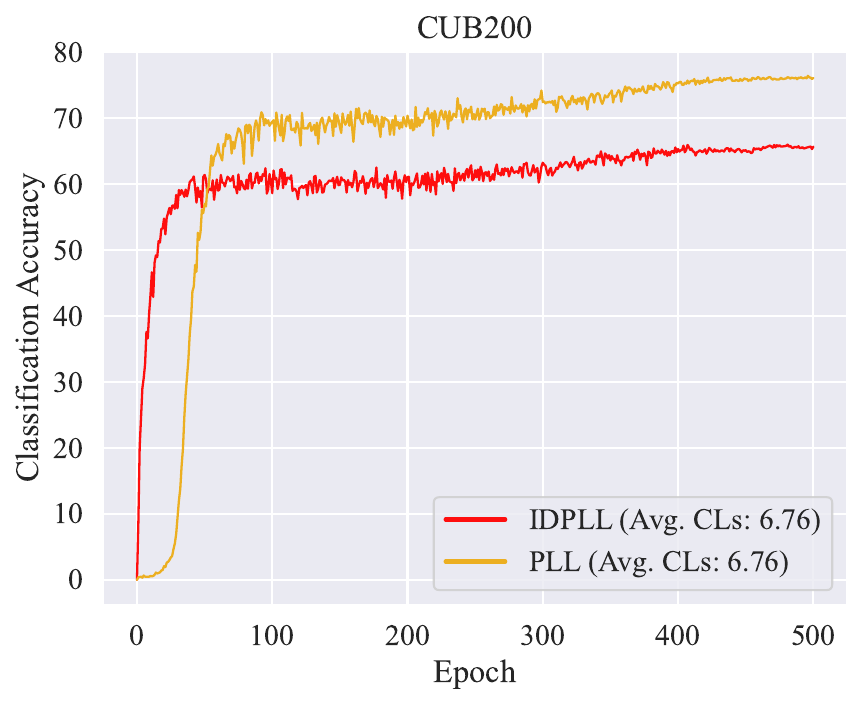}}
  \subfigure[FGVC100]{
  \includegraphics[scale=0.29]{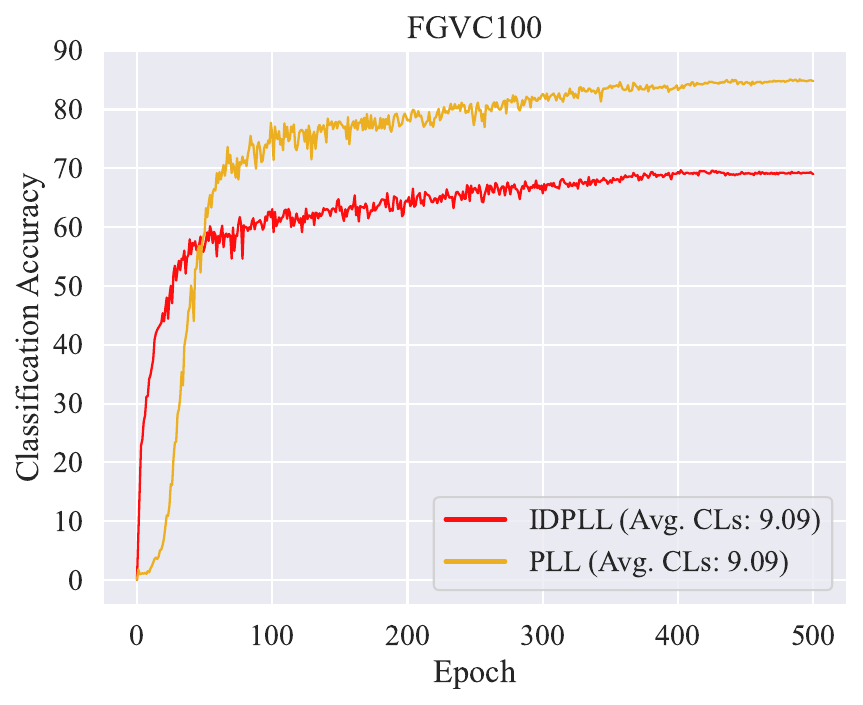}}
  \caption{The classification accuracy curves of PLL method PRODEN \cite{PRODEN} in IDPLL and PLL settings on two data sets CUB200 and FGVC100, where AVG. CLs represent the average number of candidate labels of each sample. In the IDPLL setting, the model has a faster learning speed in the early stage of training because the instance-dependent noisy labels are related to the sample to some extent, which can provide more supervision in the early stage of model training. However, in the later stage of training, the performance of PRODEN in the IDPLL setting is significantly inferior to that in the PLL setting as instance-dependent noisy labels bring more label ambiguity.}
  \label{IDPLLandPLL}
\end{figure}

IDPLL is a mixed blessing. On the positive side, as shown in Fig. \ref{IDPLLandPLL}, the model achieves better classification performance in the early stage and converges faster in the IDPLL setting compared with the PLL setting. This is because in IDPLL the noisy labels in the candidate label set are related to the sample, which can describe the sample to some extent. Although the noisy labels in the candidate label set are incorrect, they can still act as supervision to promote model training. On the negative side, the model only achieves quite inferior classification performance in the later stage. The reason is that given the instance, the instance-dependent noisy labels in the candidate label set are highly similar to the ground-truth label, bringing more label ambiguity and making the candidate label disambiguation process more challenging. The current IDPLL methods have made different attempts to achieve better label disambiguation. For example, NEPLL \cite{NEPLL} proposed a well-disambiguated sample selection method based on normalized cross-entropy and trained the model progressively according to the selected samples. POP \cite{POP} proposed to purify the candidate label set during the training phase to gradually reduce the difficulty of label disambiguation. DIRK \cite{DIRK} proposed a label rectification strategy that ensured the model output on the candidate label set was always higher than that on the non-candidate label set. However, these methods did not comprehensively exploit both the positive side and negative side of IDPLL, limiting their performance.

To address this double-edged sword challenge in IDPLL, we propose a novel method called \textbf{CEL} (Mixed Blessing: \textbf{C}lass-Wise \textbf{E}mbedding guided Instance-Dependent Partial \textbf{L}abel Learning). Specifically, unlike previous PLL and IDPLL methods, where each sample corresponds to a single embedding, our method introduces the class-wise embedding for each sample, i.e., each sample has multiple embeddings corresponding to different classes, to comprehensively exploit the mixed blessing in IDPLL. First, we propose a class associative loss (CAL) to leverage the relationships among different classes of each sample to guide the learning of class-wise embeddings. In IDPLL, the class-wise embeddings within the candidate label set should exhibit high similarity to each other as the candidate labels can describe the instance to some extent. In contrast, the class-wise embeddings between the candidate label set and the non-candidate label set should display stark differences. In this way, we can obtain class-wise embeddings that are more suitable for IDPLL. Second, we propose a prototype discriminative loss (PDL) by constructing class prototype for each class which containts global feature information to guide the label disambiguation process. To be specific, we select the most high-confidence label for each sample based on the model output, and then we ensure the class-wise embedding of this particular high-confidence label is aligned with its corresponding class prototype while being distanced from other class prototypes, thereby enhancing the model’s discriminative ability. By employing the above two losses, we can obtain embeddings that are tailor-made for IDPLL and enhance the model's label disambiguation performance simultaneously, thus addressing the mixed blessing issue in IDPLL. Extensive experiments on 6 benchmark
data sets demonstrate the effectiveness of
the proposed method when compared with 12 state-of-the-art methods. 

The contributions of our work are summarized as follows:
\begin{itemize}[leftmargin=0.75cm]
\item To the best of our knowledge, we are the first to introduce class-wise embedding in IDPLL. Class-wise embedding enable the model to explore the nuances relationships of classes in each sample, thereby better addressing the mixed blessing problem inherent in IDPLL. 
\item We comprehensively consider the positive and negative sides of IDPLL. To leverage the positive side, we utilize the class associative loss to exploit the relationships among the labels (including both candidate labels and non-candidate labels) of each sample. To address the negative side, we apply the prototype discriminative loss to utilize the relationships between high-confidence class and class prototypes.
\item Extensive experiments on benchmark data sets demonstrate that our method achieves significantly superior performance when comparing with both PLL and IDPLL methods.
\end{itemize}

\begin{figure*}[t]
    \centering
    \includegraphics[width=1\linewidth]{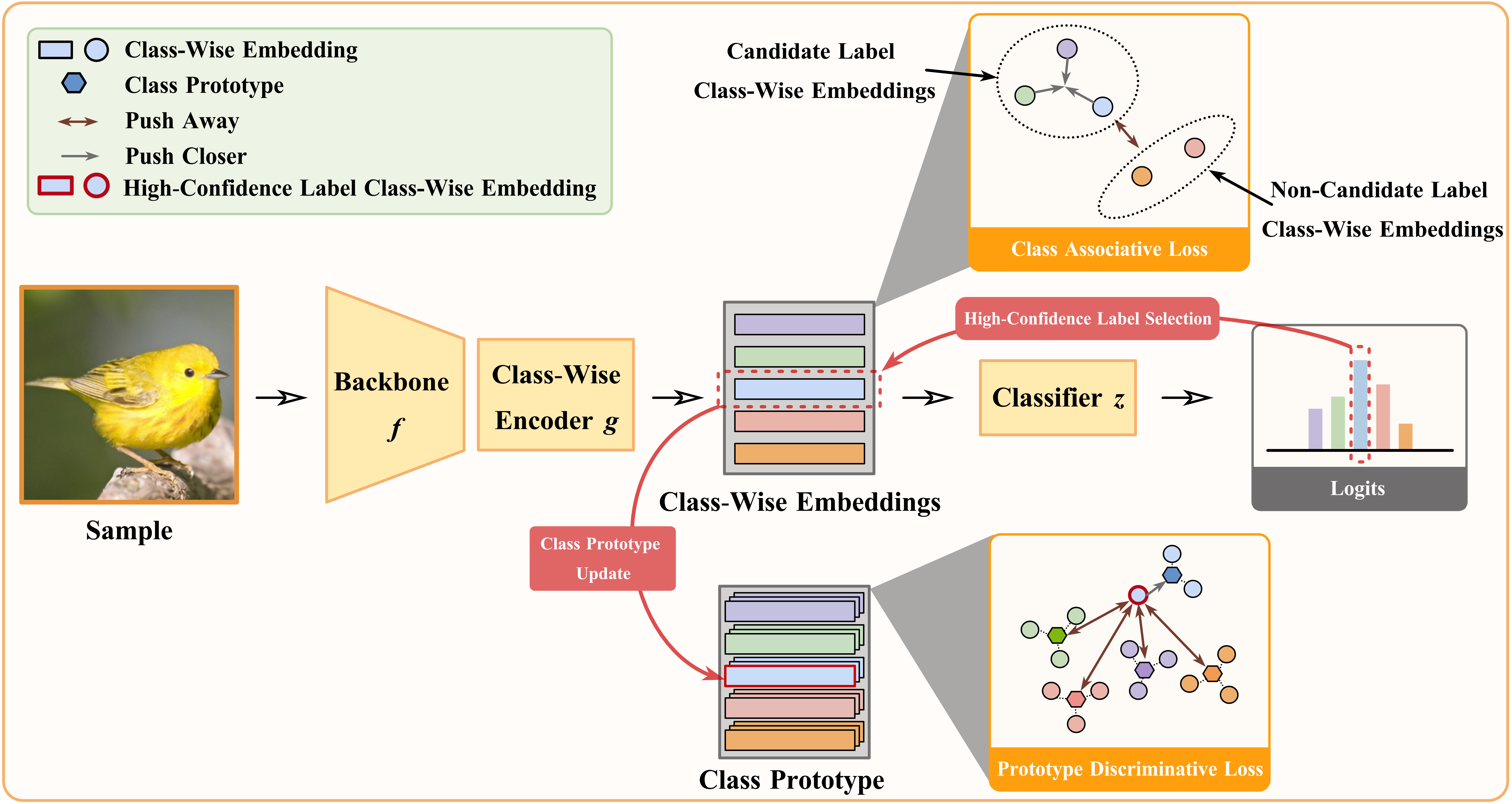}
    \caption{Illustration of our method CEL. Our model consists of three modules: the backbone $f$, the class-wise encoder $g$, and the classifier $z$. For each sample, after processing through the backbone $f$ and class-wise encoder $g$, each sample obtains its class-wise embeddings, i.e., each class corresponds to an embedding for that sample. The red line represents the process of constructing class prototypes based on the high-confidence class selected according to the model's output. The class associative loss (CAL) considers the relationships among each sample's different class-wise embeddings. While the prototype discriminative loss (PDL) considers the relationships between high-confidence class and class prototypes.}
    \label{stage}
\end{figure*}

\section{RELATED WORK}
\subsection{Partial Label Learning}
Conventional PLL methods can be roughly divided into two categories based on the used information. The first category uses the information in the feature space to guide label disambiguation \cite{PLKNN, IPAL, LEAF, LALO, AGGD}. The second category leverages the information of label space to achieve label disambiguation \cite{SDIM, PANG, DPCLS}. These methods often rely on linear models and are difficult to apply to large-scale data sets. Deep PLL has received wide attention in recent years and it adopts deep neural networks to process large-scale data sets \cite{PRODEN,DPLL,PAPI}. PRODEN \cite{PRODEN} proposed to progressively identify the ground-truth label during the self-training procedure. RC and CC \cite{RC} proposed provably risk-consistent and classifier-consistent label disambiguation methods. LWS \cite{LWS} proposed a family of leveraged weighted loss functions. PICO \cite{PICO} introduced the widely used contrastive loss \cite{MICO} to the PLL, which became the foundation for a large number of follow-up works. CR-DPLL \cite{DPLL} employed consistency regularization to reduce the impact of noisy labels. PAPI \cite{PAPI} constructed the similarity score between feature prototypes and sample embeddings, and improves the model performances by aligning the similarity score with the model output. CROSEL \cite{CroSel} used two models to cross-select trustworthy samples from the data set for the training phase. The above methods have achieved superior results in the conventional PLL setting, where the noisy labels in the candidate label set are randomly generated. However, in practice, noisy labels are always instance-dependent \cite{VALEN,ABLE,IDGP}. For example, in crowdsourced annotation tasks, the annotations from many annotators constitute the candidate label set of each sample, and the noisy labels within it are all instance-dependent. Consequently, the performances of conventional PLL methods are often limited due to the lack of consideration of the characteristics of IDPLL.

\subsection{Instance-Dependent Partial Label Learning}
IDPLL is a PLL framework that is closer to real-world scenarios. VALEN \cite{VALEN} was the first work to introduce the concept of IDPLL, with a two-stage disambiguation process. The first stage involves recovering samples' latent label distribution, and then training the model using the recovered label distribution in the second stage. ABLE \cite{ABLE} proposed an ambiguity-induced positive selection contrastive learning framework for IDPLL to achieve label disambiguation. NEPLL \cite{NEPLL} introduced a sample selection method based on normalized entropy, intending to separate well-disambiguated samples and under-disambiguated samples. It also established a dynamic candidate-aware thresholding for the further refinement of sample selection. POP \cite{POP} presented a method that progressively purified the candidate label set and optimized the classifier. IDGP \cite{IDGP} modeled the candidate label generation process for IDPLL, simulating the ground-truth label and noisy labels generation processes with Categorical distribution and Bernoulli distribution respectively. DIRK \cite{DIRK} proposed a self-distillation-based label disambiguation method, i.e., the training of the student model is directed by the output of the teacher model. It also developed a label confidence rectification method that meets the prior knowledge of IDPLL. However, the aforementioned IDPLL methods did not consider that the IDPLL setting is a mixed blessing, i.e., on the positive side, noisy labels can describe the sample to some extent and help the model training, while on the negative side, identifying the ground-truth label within the candidate label set becomes more challenging. They only focused on developing better disambiguation methods while ignoring the noisy label information, thereby limiting their performance.

\section{Proposed Method}
\subsection{Class-Wise Embedding}
Denote by $\mathcal{X} \subset \mathbb{R}^d$ the $d$-dimensional feature space and $\mathcal{Y}=\{1,2,...,q\} $ the corresponding label space with $q$ classes. Let $\mathcal{D}=\{(x_i,~S_i),1\leq i\leq m\}$ denote a partial label data set comprising $m$ samples,  where $S_i$ is the candidate label set of sample $x_i$ and the ground-truth label of sample is concealed in $S_i$. 
The objective of IDPLL is to induce a multi-class classifier that maps elements from $\mathcal{X}$ to $\mathcal{Y}$.
Previous models in the realm of PLL and IDPLL typically consist of two modules.
The first module is the backbone responsible for deriving the feature map $M\in\mathbb{R}^{H \times W \times C}$ for each sample, where $H$, $W$ and $C$ are the height, width, and channel respectively. Afterward, $M$ is processed through average global pooling to obtain the low dimensional embedding. The second module encompasses a linear layer that translates this low-dimensional embedding into the final prediction.

In this paper, we use a different paradigm that consists of three modules. As shown in Fig. \ref{stage}, the first module is as same as other models, i.e., a backbone $f$, which extracts the feature map $M_i=f(x_i)\in\mathbb{R}^{H \times W \times C}$ of each sample $x_i$ through the deep neural network. The second module is a 
\textbf{\textit{class-wise embedding}} encoder \cite{CWE1, CWE2, MLdecoder} $g$ which produces class-wise embeddings $E_i=g(M_i)\in\mathbb{R}^{q\times l}$, where $l$ is the length of each class-wise embedding. Note, the $E_i^j$ indicates the representation of the $j$-th class of sample $x_i$. The third module is a group of linear layers $z$ that output the predicted probabilities $P_i=z(E_i)\in \mathbb{R}^q$. 

In conventional PLL and IDPLL representation methods, the features of each sample are extracted as a single embedding, so they can only consider relationships at the sample level. However, in IDPLL, the relationships between each sample's candidate labels and non-candidate labels are valuable and worth leveraging. By employing class-wise encoder, we can represent each sample's embeddings on different labels. This allows us to explore the internal relationships among different classes and fully utilize the prior knowledge of IDPLL, making it a more suitable representation method for IDPLL.

\subsection{Label Disambiguation Loss}
As aforementioned, the pivotal process in addressing PLL is label disambiguation, which mitigates the impact of noisy labels within the candidate label set.  Here, we adopt a widely used deep PLL disambiguation strategy \cite{PRODEN}, which constructs sample label confidences based on model outputs during training. Initially, the label confidence vector $T_i \in \mathbb{R}^q$ of sample $x_i$ is initialized as $T_{ij}=\frac{1}{|S_i|}$, if $j \in S_i$ and $T_{ij}=0$, otherwise, where $|S_i|$ returns the number of candidate labels of sample $x_i$. Throughout the training, we update the label confidence according to the model output:
\begin{equation}
\label{T}
T_{i j}=\left\{
\begin{aligned}
&\frac{P_i^j}{\sum_{k\in S_i}P_i^k}, \qquad \text{if} \; j\in S_i, \\
&\qquad 0, \qquad \quad \;\; \text{otherwise,} 
\end{aligned}
\right.
\end{equation}
where $P_i^j$ represents the model prediction on $j$-th class of sample $x_i$. The goal of Eq. (\ref{T}) is to eliminate the influence of non-candidate labels, so that the model can focus on the candidate labels. Based on the disambiguated confidence, we can obtain the following classification loss to guide the model training:
\begin{equation}
\label{CLS}
\mathcal{L}_{cls}=\frac{1}{m}\sum_{i=1}^m \ell(P_i,T_i),
\end{equation}
where $\ell$ indicates the cross-entropy loss. Eq. \eqref{CLS} is usually viewed as a self-learning process and has demonstrated efficacy across various PLL scenarios.

\subsection{Class Associative Loss based on Class-Wise Embedding}
\label{classsection}
As previous analyzed, IDPLL is a double-edged sword. On the positive side, the noisy labels in the candidate label set are very similar to the ground-truth label because they often share common features and they can depict the sample to some extent, which is also the fundamental reason leading to label ambiguity. 
Therefore, an important characteristic of IDPLL is that the labels within the candidate label set should be very similar, while the labels in the candidate label set should exhibit significant differences from the labels in the non-candidate label set. To fully leverage this prior knowledge, we can measure the relationships among classes of each sample through class-wise embeddings. To be specific, the class-wise embeddings corresponding to each label in the candidate label set should be similar to each other, while the class-wise embeddings between the candidate label set and the non-candidate label set should display stark differences. Therefore, we can construct the following class-wise relationships:
\begin{equation}
\label{s_1}
s_i^{cal}=\sum_{j,k\in S_i}\langle E_i^j,E_i^k \rangle / \sum_{j,k\in S_i} 1,
\end{equation}    

\begin{equation}
\label{d_1}
d_i^{cal}=\sum_{j \in S_i, h \notin S_i}\langle E_i^j,E_i^h \rangle / \sum_{j \in S_i, h \notin S_i} 1,
\end{equation}
\noindent where $E_i^j$ and $S_i$ represent the $j$-th class-wise embedding and the candidate label set of sample $x_i$ respectively. $\langle \cdot , \cdot \rangle$ returns the cosine similarity of two vectors. Note that the class-wise embeddings have been normalized before calculating the cosine similarity. $s_i^{cal}$ in Eq. (\ref{s_1}) denotes the average similarity of classes in the candidate label set, while $d_i^{cal}$ in Eq. (\ref{d_1}) indicates the average similarity of classes between the candidate label set and the non-candidate label set.

By considering the similarities of Eqs. (\ref{s_1}) and (\ref{d_1}) simultaneously, we can obtain the following class associative loss (CAL) function:
\begin{equation}
\label{slevel}
\mathcal{L}_{cal}=\frac{1}{m}\sum_{i=1}^m\left(\left(1-s_i^{cal}\right)+\gamma_1\left|d_i^{cal}\right|\right),
\end{equation}
\noindent where $\gamma_1$ is a trade-off parameter that balances two different terms. $|\cdot|$ is the absolute value operator given $d_i^{cal} \in (-1,1)$. The first term in Eq. (\ref{slevel}) means that for each sample $x_i$, the class-wise embeddings in its candidate label set should be pulled as close as possible. In the meanwhile, the second term implies that the class-wise embeddings between the candidate label set and the non-candidate label set should be pushed as far away as possible, and in the ideal situation, their class-wise embeddings should be orthogonal, i.e. $d_i^{cal}=0$. By minimizing the loss function $\mathcal{L}_{cal}$, we can obtain a model that is more suitable for the IDPLL setting.

\subsection{Prototype Discriminative Loss based on Class-Wise Embedding}
\label{prototypesection}
In IDPLL, we can distinguish the labels between the candidate label set and the non-candidate label set easily, however, it becomes more difficult to identify which label in the candidate label set is the ground-truth label, as the candidate labels are similar to each other, bringing more label ambiguity. Therefore, to tackle this negative side of IDPLL and identify the ground-truth label, we use the global information of samples to guide the model's disambiguation. Therefore, we first construct prototypes for each class,
\begin{equation}
\label{prototype}
Q^c=Normalize(Q^c+E_i^c), \textit{if} \;\; c=argmax (P_i)  \;\textit{and}\; c\in S_i,
\end{equation}
\noindent where $Q^c$ is the $c$-th class prototypes and $Normalize(\cdot)$ denotes the $L_2$ normalization operator. To ensure the quality of the class prototypes, we only select the class with the highest model output probability in the candidate label set as the high-confidence label of each sample and add the corresponding class-wise embedding into the class prototype.

Similar to Eq. (\ref{s_1}) and Eq. (\ref{d_1}), we construct the similarity relationships between class-wise embeddings and class prototypes as follows:
\begin{equation}
\label{s_2}
s_i^{pdl}=\langle E_i^c,Q^c\rangle, \textit{if} \;\; c=argmax(P_i)  \;\textit{and}\; c\in S_i,
\end{equation}
\begin{equation}
\label{d_2}
d_i^{pdl}=\sum_{k\neq c, k\in [q]}\langle E_i^c,Q^k\rangle / (q-1),
\end{equation}
\noindent where $s_i^{pdl}$ in Eq. (\ref{s_2}) represents the similarity between the class-wise embedding of the class with the highest model prediction in the candidate label set and the corresponding class prototype. Meanwhile, $d_i^{pdl}$ in Eq. (\ref{d_2}) denotes the average similarity between this selected class-wise embedding and all other class prototypes. By combing Eq. (\ref{s_2}) and Eq. (\ref{d_2}), we have the following prototype discriminative loss (PDL) function:
\begin{equation}
\label{plevel}
\mathcal{L}_{pdl}=\frac{1}{m}\sum_{i=1}^m\left(\left(1-s_i^{pdl}\right)+\gamma_2\left|d_i^{pdl}\right|\right),
\end{equation}
where $\gamma_2$ is the trade-off parameter that balances the two different prototype level terms. By minimizing Eq. (\ref{plevel}), we can guide the model training through class prototypes, i.e., the most reliable class predicted by the model should be as close as possible to the corresponding class prototype, while this reliable class should be far away from other class prototypes, and in the most ideal case, their similarity relationship should be 0. By utilizing the global information of class prototypes, we can effectively improve the discriminative performance of the model and select the ground-truth label from the candidate label set.

\begin{algorithm}[t]
\label{pseudo}
\caption{The Pseudo Code of the Proposed Method}
\KwIn{\\$\mathcal{D}$: the instance-dependent partial label training set;
\\$T_{max}$: the total training epoch;
\\$T_{w}$: the first stage epoch;
\\$\alpha, \beta, \gamma_1 ,\gamma_2$: the parameters of the loss function;}
\KwOut{\\model parameters.} 

\For{$t = 1$ to $T_{max}$}{

Sample a mini-batch from $\mathcal{D}$;

\eIf{$t \leq T_{w}$}{Calculate the loss according to Eq. (\ref{LW});}{Calculate the loss according to Eq. (\ref{LA});}

Update the label confidence according to Eq. (\ref{T});

Update the class prototype according to Eq. (\ref{prototype});

Update the model parameters;
}

Return result.

\end{algorithm}

\begin{table*}[ht]
  \renewcommand{\arraystretch}{1.3}
  \centering
  \caption{Characteristic of the benchmark data sets, where Avg. CLs represent the average number of candidate labels per
  sample.}
  
  \begin{tabular}{c c c c c c c c}
  \Xhline{1px}
  &Data set &Train  &Test  &Dimensions &Classes &Avg. CLs\\
  \Xhline{.5px}
  \multirow{2}{*}{Common} &CIFAR-100 &50000 &10000 &32$\times$32 &100 &10.82 (rate=0.1)\\
  &CIFAR-100H &50000 &10000 &32$\times$32 &100 &3.41 (rate=0.6)\\
  
  \Xhline{0.5px}
  \multirow{4}{*}{Fine-Grained} 
  &FGVC Aircraft (FGVC100) &6776 &3333 &224$\times$224 &100 &9.09 (rate=0.1)\\
  &Stanford Dogs (DOGS120) &12000 &8580 &224$\times$224 &120 &11.51 (rate=0.1)\\
  &Stanford Cars (CARS196) &8144 &8041 &224$\times$224 &196 &9.98 (rate=0.05)\\
  &CUB-200-2011 (CUB200) &5994 &5794 &224$\times$224 &200 &6.76 (rate=0.03)\\
  \Xhline{1px}
  \end{tabular}
  \label{data}
  \end{table*}

  \begin{table*}[ht]
      \renewcommand{\arraystretch}{1.4}
      \centering
      \caption{Classification accuracy (mean±std) of each method on benchmark data sets. Bold and underlined indicate the best and second-best results, respectively.}
      \label{classification}
      \begin{tabular}{c c c c c c c c}
          \Xhline{.5px}
          \Xhline{.5px} 
          Type &Method &CIFAR-100 & CIFAR-100H & FGVC100 & DOGS120 & CARS196 & CUB200\\ 
          \Xhline{.5px} 
          \multirow{7}{*}{IDPLL}  &\cellcolor{gray!25} CEL (OURS) 
          &\cellcolor{gray!25}\textBF{75.51±0.28\%} &\cellcolor{gray!25}\textBF{75.77±0.09\%} &\cellcolor{gray!25}\textBF{78.36±0.19\%}  &\cellcolor{gray!25}\textBF{78.18±0.12\%} &\cellcolor{gray!25}\textBF{86.22±0.08\%} &\cellcolor{gray!25}\textBF{68.60±0.10\%}\\ 
          &DIRK (AAAI 2024) &\underline{74.40±0.18\%} &\underline{75.50±0.45\%}  &76.86±3.54\%  &\underline{75.97±0.29\%} &85.31±0.77\% &\underline{66.60±1.07\%}\\ 
          &NEPLL (ICCV 2023) &72.41±0.66\% &75.05±0.83\% &75.36±0.59\% &74.84±0.09\% &85.05±0.17\% &62.88±1.66\% \\ 
          &POP (ICML 2023) &72.74±0.70\% &75.11±0.18\%   &\underline{77.87±0.23\%} &74.86±0.12\%  &85.26±0.24\% &64.88±0.48\%\\ 
          &IDGP (ICLR 2023) &68.49±0.35\% &68.83±1.11\%  &72.48±0.86\% &66.79±0.38\% &79.56±0.46\%  &58.16±0.58\%\\ 
          &ABLE (IJCAI 2022) &70.94±0.17\% &73.15±0.15\%  &74.05±0.43\% &72.78±0.07\% &\underline{85.75±0.24\%} &63.23±0.36\%\\ 
          &VALEN (NeurIPS 2021) &68.33±0.36\% &70.52±0.24\%  &68.21±0.43\%  &66.89±0.15\% &82.21±0.16\% &63.05±3.32\%\\ 
          \Xhline{.5px} 
          \multirow{6}{*}{PLL} &PICO (ICLR 2022) &64.00±0.29\% &65.39±0.38\% &63.52±0.94\% &67.80±-0.06\%  &70.15±1.63\% &58.56±0.90\%\\ 
          &CR-DPLL (ICML 2022) &71.54±0.25\% &75.49±0.30\%    &63.20±1.22\% &61.41±0.82\% &68.30±0.45\% &50.26±0.34\% \\ 
          &LWS (ICML 2021) &71.64±0.32\% &74.40±0.62\%   &72.82±0.44\% &66.12±0.12\%  &82.11±0.24\%  &54.01±0.68\%\\ 
          &PRODEN (ICML 2020) &70.49±0.50\% &72.90±0.47\%    &69.34±0.39\% & 70.94±0.43\% &83.35±0.05\% &65.06±0.14\%\\ 
          &RC (NeurIPS 2020) &69.96±0.01\%  &72.69±0.30\%  &72.72±0.22\%  &70.40±0.26\% &82.11±0.63\% &60.96±0.59\%\\ 
          &CC (NeurIPS 2020) &70.51±0.28\% &73.83±0.19\%   &64.36±0.50\% &66.21±0.77\%  &70.61±2.01\% &56.79±0.71\%\\ 
          \Xhline{.5px}
          \Xhline{.5px} 
      \end{tabular}
  \end{table*}

\subsection{Overall Objective}
Considering the low quality of class prototypes obtained in the early stages of model training, it is difficult to ensure the effectiveness of label disambiguation. Therefore, we divide the model training into two stages. In the first stage, our training uses the classification loss and class associative loss which aims to learn a model more suitable for IDPLL. The training objective of the first stage can be written as follows:
\begin{equation}
\label{LW}
\mathcal{L}_{all}=\mathcal{L}_{cls}+\alpha\mathcal{L}_{cal}.
\end{equation}
After training for $T_w$ epochs, we add the prototype discriminative loss into the training objective to further improve the model's disambiguation performance. The objective function of the second stage can be summarized as follows: 
\begin{equation}
\label{LA}
\mathcal{L}_{all}=\mathcal{L}_{cls}+\alpha\mathcal{L}_{cal}+\beta\mathcal{L}_{pdl},
\end{equation}
\noindent where $\alpha$ and $\beta$ are two trade-off parameters. The overall pseudo-code of our method is summarized in \textbf{Algorithm 1}.

\section{EXPERIMENTS}
\subsection{Experimental Setting}
\subsubsection{Data sets}
To demonstrate the effectiveness of our method, we conducted comparisons on several challenging data sets with at least 100 classes. To be specific, we conducted experiments on two common image data sets including CIFAR-100 \cite{cifar}, CIFAR-100H \cite{cifar} and four fine-grained \cite{finegrained,finegrained1} image data sets including CUB-200-2011 \cite{cub}, Stanford Cars \cite{car}, FGVC Aircraft \cite{fgvc}, Stanford Dogs \cite{dog} which are more easily to cause label ambiguity. Table \ref{data} records the detailed information of all the data sets, where Avg. CLs represent the average number of candidate labels per sample. For data sets CIFAR-100 and CIFAR-100H, the image size was set to 32 $\times$ 32, while for fine-grained data sets, we resized the images to 224 $\times$ 224. We employed the IDPLL noisy label generation method proposed by VALEN \cite{VALEN} to generate instance-dependent noisy labels. Note that for CIFAR-100H, noisy labels only appear in other subclasses that belong to the same superclass as the ground-truth label.

\begin{table*}[t]
  \caption{Win/tie/loss counts on the classification performance of our method CEL against each comparing method on benchmark
  data sets according to the pairwise t-test at 0.05 significance level.}
      \renewcommand{\arraystretch}{1.3}
      \centering
      \label{ttest}
      \begin{tabular}{c c c c c c c c c c c c c c}
          \Xhline{.5px}
          \Xhline{.5px} 
          &DIRK &NEPLL &POP &IDGP &ABLE &VALEN & PICO & CR-DPLL & LWS & PRODEN &RC &CC &\textBF{Total}\\
          \Xhline{.5px} 
           Common &1/1/0 &1/1/0 &1/1/0 &2/0/0 &2/0/0 &2/0/0 &2/0/0 &1/1/0 &2/0/0 &2/0/0 &2/0/0 &2/0/0 &20/4/0\\ 
           Fine-Grained &2/2/0 &4/0/0 &4/0/0 &4/0/0 &4/0/0 &4/0/0 &4/0/0 &4/0/0 &4/0/0 &4/0/0 &4/0/0 &4/0/0 &46/2/0\\
           \rowcolor{gray!25}Total &3/3/0 &5/1/0 &5/1/0 &6/0/0 &6/0/0 &6/0/0 &6/0/0 &5/1/0 &6/0/0 &6/0/0 &6/0/0 &6/0/0 &66/6/0\\
          \Xhline{.5px}
          \Xhline{.5px} 
      \end{tabular}
  \end{table*}
  
  \begin{figure}[t]
  \centering
  \includegraphics[width=0.8\linewidth]{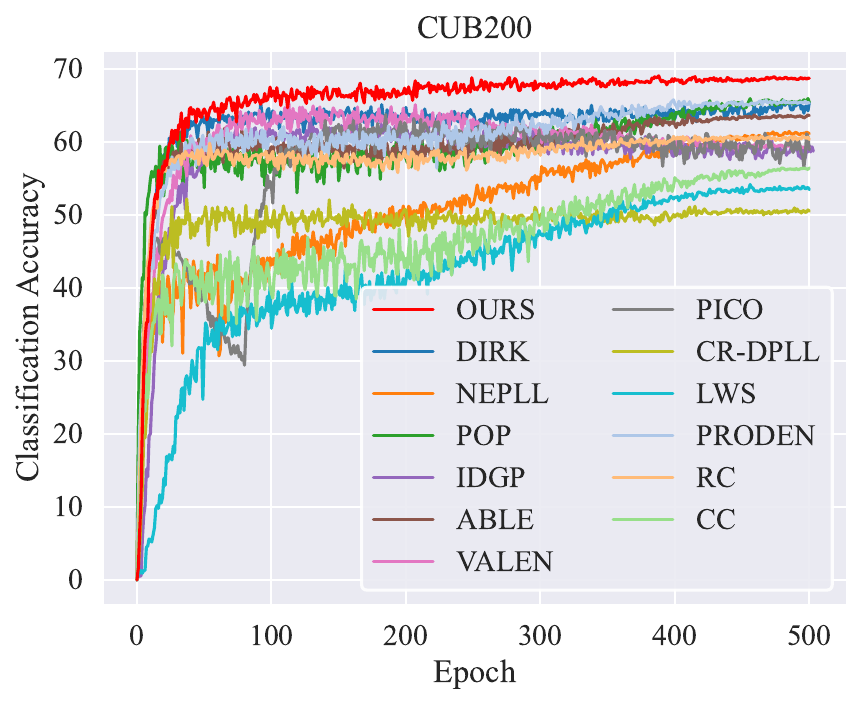}
  \caption{Classification accuracy curves of all methods on benchmark data set CUB200.}
  \label{converage}
  \end{figure}

\subsubsection{Comparing methods}
To demonstrate the effectiveness of the proposed method, we compared our method with 12 methods including 6 IDPLL methods and 6 PLL methods. \textit{IDPLL methods}: (i) DIRK \cite{DIRK}, a self-distillation based label disambiguation method. (ii) NEPLL \cite{NEPLL}, a normalized entropy guided sample selection method. (iii) POP \cite{POP}, a method that progressive purifies candidate label set and refines classifier. (iv) IDGP \cite{IDGP}, a generation method that models the candidate label generation process. (v) ABLE \cite{ABLE}, a contrastive learning-based framework that uses ambiguity-induced positives selection method. (vi) VALEN \cite{VALEN}, a label enhancement guided latent label distribution recovery method. \textit{PLL methods}:  (i) PICO \cite{PICO}, a method that combines PLL and contrastive learning for the first time. (ii) CR-DPLL \cite{DPLL}, a consistency regularization label disambiguation method. (iii) LWS \cite{LWS}, a method that uses leveraged weighted loss to balance candidate label set and non-candidate label set. (iv) PRODEN \cite{PRODEN}, a method that progressively identifies the
ground-truth label during the self-training procedure. (v) RC \cite{RC}, a risk-consistent weighting method. (vi) CC \cite{RC}, a classifier-consistent that uses a transition matrix. The parameters of all methods were set according to their original papers.

\subsubsection{Implementation details}
For fair comparisons, we employed ResNet-18 \cite{resnet} as the backbone $f$ on data sets CIFAR-100 and CIFAR-100H, while using a ResNet-34 \cite{resnet} pre-trained on ImageNet \cite{imagenet} as the backbone $f$ on fine-grained data sets for all methods. We used ML-Decoder \cite{MLdecoder} as our class-wise encoder $g$. Stochastic gradient descent (SGD) was used as the optimizer with a momentum of 0.9 and all methods were trained for 500 epochs. We selected learning rate in \{0.5, 0.1, 0.05, 0.01, 0.005, 0.001\} and weight decay in \{0.001, 0.0005, 0.0001\} respectively. Additionally, we applied the cosine annealing learning rate schedule for all methods. The batch sizes of all data sets were set to 128. We repeated the experiments three times under the same random seeds and recorded the mean accuracy and standard deviation. As for our method, parameters $\alpha$ and $\beta$ were selected from $\{0.1,0.5,1\}$, while $\gamma_1$ and $\gamma_2$ were selected from $\{0.5,1,2,5\}$. The $T_w$ was set to 250, which is half of the total training epochs. The length of each class-wise embedding $l$ was set to 512 on all data sets. Following DIRK \cite{DIRK}, we used the same weak augmentation and strong augmentation for our method.

\subsection{Experimental Results}
\subsubsection{Classification performance}
Table \ref{classification} reports the classification accuracies of all methods on two common data sets and four fine-grained data sets. According to Table \ref{classification}, our method ranks first in all benchmark data sets when compared with 6 PLL methods and 6 IDPLL methods. It is worth noting that there is only minor difference in classification accuracies between the previous PLL and IDPLL methods on common data sets like CIFAR-100 and CIFAR-100H. While our method improves classification accuracy from 74.40±0.18\% to 75.51±0.28\% on the CIFAR-100 data set compared to the best previous method. On fine-grained data sets, the performance gaps between PLL and IDPLL methods are significant. PLL methods generally fail to achieve satisfactory accuracy because all labels in the fine-grained data sets belong to the same superclass, which is more challenging. Consequently, IDPLL methods tailored to this setting often outperform the conventional PLL methods. Our method CEL constantly excels on these fine-grained data sets, improving classification accuracy from 66.60±1.07\% to 68.60±0.10\% on the data set CUB200 and from 75.97±0.29\% to 78.18±0.12\% on the data set DOGS120, compared to the best previous method.

\subsubsection{Significance test}
Table \ref{ttest} reports win/tie/loss counts of our method CEL against each comparing method based on the pairwise t-test at 0.05 significance level. As shown in Table \ref{ttest}, on common data sets, our method CEL significantly outperforms the comparing methods in 83.3\% (20/24). While on fine-grained data sets, our method significantly outperforms the comparing methods in 95.8\% (46/48). Furthermore, our method achieves superior performance than all comparing methods except DIRK on fine-grained data sets. Considering all benchmark data sets, our method significantly outperformed the comparing methods in 91.7\% (66/72), demonstrating the effectiveness of our method.

\subsubsection{Classification accuracy curves}
Fig. \ref{converage} shows the classification accuracy curves of all methods on data set CUB200. As shown in Fig. \ref{converage}, compared to other methods, our method CEL maintains a relatively fast learning speed in the early stages, only slightly slower than POP, demonstrating that our class associative loss can enhance the model's learning speed. Moreover, CEL achieves the best classification accuracy in the later stages of training, proving that our prototype discriminative loss further improves the model's disambiguation performance.

\begin{table*}[ht]
\caption{Ablation study of our method on the benchmark data sets, where $\mathcal{L}_{cls}$, $\mathcal{L}_{cal}$ and $\mathcal{L}_{pdl}$ indicate the label disambiguation loss, the class associative loss and the prototype discriminative loss respectively.}
\label{ablation}
    \renewcommand{\arraystretch}{1.3}
    \centering
    \begin{tabular}{c c c c c c c c c c}
        \Xhline{.5px}
        \Xhline{.5px} 
        $\mathcal{L}_{cls}$ &$\mathcal{L}_{cal}$ &$\mathcal{L}_{pdl}$ &CIFAR-100 & CIFAR-100H  & FGVC100 & DOGS120 & CARS196 & CUB200 &\textBF{AVERAGE}\\
        \Xhline{.5px} 
        \Checkmark & &  &73.70±0.08\% &75.21±0.11\% &76.65±0.33\%  &74.83±0.75\%  &84.69±0.87\% &65.99±0.99\% &75.18\%\\ 
        \Checkmark &\Checkmark &  &74.68±0.12\% &75.61±0.06\%&77.71±0.17\%  &77.33±1.03\% &85.74±0.10\% &67.32±1.04\% &76.40\%\\ 
        \rowcolor{gray!25}\Checkmark &\Checkmark &\Checkmark &\textBF{75.51±0.28\%}  &\textBF{75.77±0.09\%}  &\textBF{78.36±0.19\%} &\textBF{78.18±0.12\%}  &\textBF{86.22±0.08\%}  &\textBF{68.60±0.10\%} &\textBF{77.11\%} \\ 
        \Xhline{.5px}
        \Xhline{.5px} 
    \end{tabular}
\end{table*}

\begin{figure*}[ht]
    \centering
    \includegraphics[width=1\linewidth]{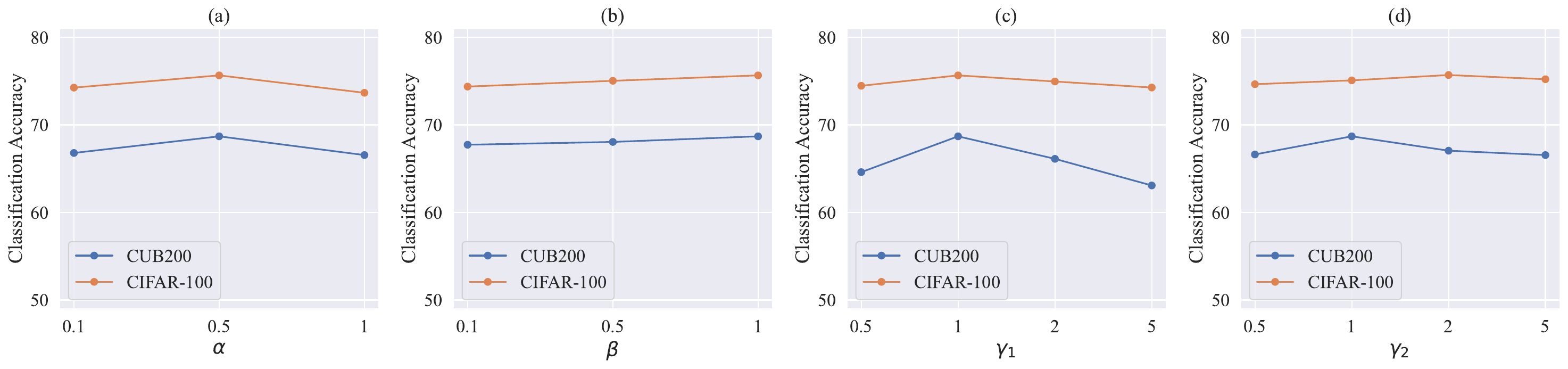}
    \caption{Parameters sensitivity of our method CEL. (a) - (d) represent the classification accuracy of our method on benchmark data sets CIFAR-100 and CUB200 by varying $\alpha$, $\beta$, $\gamma_1$ and $\gamma_2$ respectively.}
    \label{sensitive}
\end{figure*}

\subsection{Further Analysis}
\subsubsection{Ablation study}
In Table \ref{ablation}, we conduct ablation studies on all benchmark data sets to demonstrate the necessity of each component in our method. The first row represents our method with only a classification loss $\mathcal{L}_{cls}$, the second row represents our method with the classification loss and our proposed class associative loss, i.e., $\mathcal{L}_{cls}+\mathcal{L}_{cal}$, and the third row represents our method with the classification loss, class associative loss, and our prototype discriminative loss, i.e., $\mathcal{L}_{cls}+\mathcal{L}_{cal}+\mathcal{L}_{pdl}$. According to the results in Table \ref{ablation}, both the class associative loss $\mathcal{L}_{cal}$ and prototype discriminative loss $\mathcal{L}_{pdl}$ can improve the model's classification performance. To be specific, the class associative loss improves the classification accuracy by an average of 1.22\% on six data sets, and the prototype discriminative loss $\mathcal{L}_{pdl}$ loss also promotes the classification accuracy by an average of 0.71\%. Therefore, incorporating both terms into the model is the optimal choice.
\subsubsection{Parameters sensitivity}
Fig. \ref{sensitive} shows the classification accuracy of our method CEL on benchmark data sets CIFAR-100 and CUB200 under different parameter settings. Fig. \ref{sensitive} (a), (b), (c) and (d) correspond to the parameters $\alpha$, $\beta$, $\gamma_1$, and $\gamma_2$, respectively. To be specific, $\alpha$ and $\beta$ control the weights of the class associative loss and the prototype discriminative loss, while $\gamma_1$ and $\gamma_2$ control the relative importance of the pull close and push away components within each loss. As illustrated in Fig. \ref{sensitive}, when $\alpha$ is set to 0.5, $\beta$ to 1, and $\gamma_1$ to 1, the model achieves the best classification performance. Specifically, when $\gamma_2$ is set to 1 and 2, our method achieves the highest classification accuracy on data sets CUB200 and CIFAR-100, respectively. Therefore, setting $\alpha$, $\beta$, and $\gamma_1$ to 0.5, 1, and 1, and choosing $\gamma_2$ from \{1, 2\} are the suggested parameters for our method.

\begin{figure}[t]
    \centering
    \includegraphics[width=0.8\linewidth]{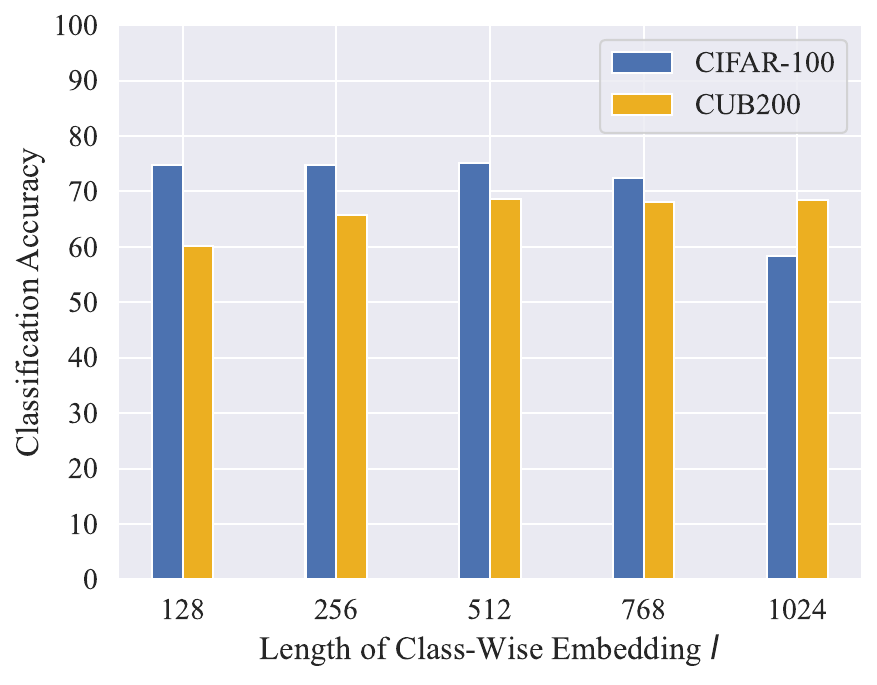}
    \caption{Classification accuracy of different lengths of class-wise embedding on data sets CIFAR-100 and CUB200.}
    \label{embedding}
\end{figure}

\subsubsection{Length of the class-wise embedding}
We conduct experiments on the data sets CIFAR-100 and CUB200 to verify the impact of different lengths of class-wise embedding $l$ on model classification performance. To be specific, we select $l$ in $\{128,256,512,768,1024\}$. As shown in Fig. \ref{embedding}, for data set CIFAR-100, which has smaller image sizes ($32\times32$), the classification accuracy of the model is higher when the length of the class-wise embedding is less than or equal to 512. This is because excessively large embeddings can dilute important features in data sets with smaller feature dimensions, which reduces classification performance. Conversely, for the data set CUB200, which has larger image sizes ($224\times224$), the classification accuracy is higher when the length of the class-wise embedding is greater than or equal to 512, this is because for data sets with larger feature dimensions, excessively small embeddings can compress important feature information, leading to a decline in performance. Therefore, considering both cases, setting the class-wise embedding length to 512 is a good choice.
\section{CONCLUSION}
In this paper, we have presented a novel method named CEL to address the IDPLL problem. For the first time, we realize that IDPLL is a mixed blessing with both positive side and negative side.  We, therefore, propose to construct the class-wise embeddings to explore the relationships among the candidate labels and the non-candidate labels. To leverage the positive side of IDPLL, we introduced the class associative loss to learn representations that are more suitable for IDPLL. This is achieved by leveraging the similarity among labels within the candidate label set and the differences between the candidate label set and the non-candidate label set through class-wise embeddings. To mitigate the negative side of IDPLL, i.e., identifying the ground-truth label in the candidate set becomes more challenging, we constructed prototype discriminative loss to guide the model’s disambiguation process using class prototypes which include global sample information. Extensive experiments on both common and fine-grained data sets demonstrate that our method significantly outperforms twelve state-of-the-art PLL and IDPLL methods. Moreover, compared with previous methods, our method converges faster in the early stages of model training, while produces the highest classification accuracy in the later training stages.
\clearpage
\newpage

\balance
\bibliographystyle{ACM-Reference-Format}
\bibliography{KDD25}

\setcounter{equation}{0}
\setcounter{table}{0}
\appendix
\clearpage

\renewcommand\thesection{\Alph{section}}
\renewcommand*{\thetable}{S\arabic{table}}
\renewcommand*{\thefigure}{S\arabic{figure}}
\renewcommand*{\theequation}{S\arabic{equation}}

\end{document}